\documentclass{article}

\usepackage{microtype}
\usepackage{graphicx}
\usepackage{subfigure}
\usepackage{booktabs} 

\usepackage{amsmath,amsfonts,bm}

\def\eqref#1{equation~\ref{#1}}

\def\1{\bm{1}}

\def\veps{{\bm{\epsilon}}}

\def\vc{{\bm{c}}}
\def\vd{{\bm{d}}}

\def\vr{{\bm{r}}}

\def\vx{{\bm{x}}}

\def\mI{{\bm{I}}}

\def\mZ{{\bm{Z}}}

\DeclareMathAlphabet{\mathsfit}{\encodingdefault}{\sfdefault}{m}{sl}
\SetMathAlphabet{\mathsfit}{bold}{\encodingdefault}{\sfdefault}{bx}{n}

\newcommand{\proj}{\mathcal{P}}

\usepackage{epsfig}
\usepackage{graphicx}
\usepackage{wrapfig}
\usepackage{tabularx}
\usepackage{tikz}
\usepackage{enumitem}
\usepackage{comment}
\usepackage{bm}
\usepackage{multirow}
\usepackage{booktabs}

\usepackage{hyperref}

\usepackage[linesnumbered,ruled,vlined,algo2e]{algorithm2e}
\SetCommentSty{mycommfont}
\SetKwInput{KwInput}{Input}                
\SetKwInput{KwOutput}{Output}              

\usepackage[arxiv]{icml2023}

\usepackage{amsmath}
\usepackage{amssymb}
\usepackage{mathtools}
\usepackage{amsthm}

\usepackage[capitalize,noabbrev]{cleveref}
\crefformat{equation}{Eq.~(#2#1#3)}
\crefname{section}{§}{§§}
\Crefname{section}{§}{§§}

\theoremstyle{plain}

\theoremstyle{definition}

\theoremstyle{remark}

\newcommand{\model}{NerfDiff}
\newcommand{\pixelnerf}{single-image NeRF}
\newcommand*\diff{\mathop{}\!\mathrm{d}}
\usepackage[textsize=tiny]{todonotes}

\icmltitlerunning{{\model}: Single-image View Synthesis with NeRF-guided Distillation from 3D-aware Diffusion}

\begin{document}

\twocolumn[
\icmltitle{{\model}: Single-image View Synthesis with NeRF-guided Distillation \\ from 3D-aware Diffusion}




\begin{icmlauthorlist}
\icmlauthor{Jiatao Gu}{apple}
\icmlauthor{Alex Trevithick}{ucsd}
\icmlauthor{Kai-En Lin}{ucsd}
\icmlauthor{Josh Susskind}{apple}
\icmlauthor{Christian Theobalt}{mpi} \\
\icmlauthor{Lingjie Liu}{mpi,penn} 
\icmlauthor{Ravi Ramamoorthi}{ucsd}
\end{icmlauthorlist}


\icmlaffiliation{apple}{Apple}
\icmlaffiliation{penn}{University of Pennsylvania}
\icmlaffiliation{ucsd}{University of California, San Diego}
\icmlaffiliation{mpi}{Max Planck Institute for Informatics, Germany}

\icmlcorrespondingauthor{Jiatao Gu}{jiatao@apple.com}
\icmlcorrespondingauthor{Lingjie Liu}{lingjie.liu@seas.upenn.edu}

\icmlkeywords{Machine Learning, ICML}

\vskip 0.3in
]



\printAffiliationsAndNotice{}  



\begin{abstract}
Novel view synthesis from a single image requires inferring occluded regions of objects and scenes whilst simultaneously maintaining semantic and physical consistency with the input. Existing approaches condition  neural radiance fields (NeRF) on local image features, projecting points to the input image plane, and aggregating 2D features to perform volume rendering. However, under severe occlusion, this projection fails to resolve uncertainty, resulting in blurry renderings that lack details. In this work, we propose {\model}, which addresses this issue by distilling the knowledge of a 3D-aware conditional diffusion model (CDM) into NeRF through synthesizing and refining a set of virtual views at test-time. We further propose a novel NeRF-guided distillation algorithm that simultaneously generates 3D consistent virtual views from the CDM samples, and finetunes the NeRF based on the improved virtual views. Our approach significantly outperforms existing NeRF-based and geometry-free approaches on challenging datasets including ShapeNet, ABO, and Clevr3D. Please see the supplementary website (\url{https://jiataogu.me/nerfdiff}) for video results.
\end{abstract}

\section{Introduction}
\label{sec:intro}
\begin{figure}[t]
    \centering
    \includegraphics[width=\linewidth]{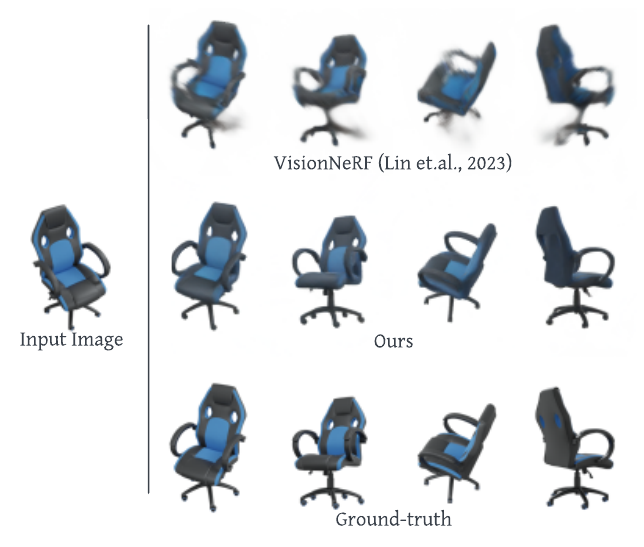}
    \vspace{-20pt}
    \caption{Renderings from our method in comparison to the SoTA VisionNeRF~\cite{visionnerf}. Note how our method can predict sharp renderings despite large occlusion, whereas VisionNeRF cannot handle this uncertainty and shows implausible blurring. } 
    \label{fig:example_nvs}
    \vspace{-10pt}
\end{figure}
Novel view synthesis is a core component of computer graphics and vision applications, including virtual and augmented reality, immersive photography, and the creation of digital replicas.
Given a few input views of an object or a scene, one seeks to synthesize
new views from other viewing directions.  
This problem is challenging since novel views must account for
occlusions and unseen regions. 
This problem has a long history, going back to early work in image-based
rendering (IBR)~\cite{ChenWilliams,Gortler,Levoy,McMillan}.  
However, IBR methods can only produce suboptimal results and are often scene-specific. 
Recently, neural radiance fields
(NeRF)~\cite{mildenhall2020nerf} have shown high-quality novel view synthesis results, but NeRF requires tens or hundreds of images for overfitting a scene and has no generalization ability to infer new scenes.

This work focuses on novel view synthesis from a single image. 
In attempts to do so, generalizable NeRF models~\cite{grf,pixelnerf,visionnerf} have been proposed, whereby the NeRF representation is conditioned by the projection of 3D points and 
gathering of corresponding image features. These approaches produce good results, especially for cameras near the input. However, when the target views are far from the input, these approaches yield blurry results. The uncertainty of large unseen regions in novel views cannot be resolved by projection to the input image. A distinct line of work addresses the uncertainty issue in single-image view synthesis by leveraging 2D generative models to predict novel views conditioned on the input view~\cite{rombach2021geometry, watson2022novel}. However, these approaches are only able to synthesize partially 3D-consistent images.

\begin{figure*}[t]
\centering
   \includegraphics[width=\linewidth]{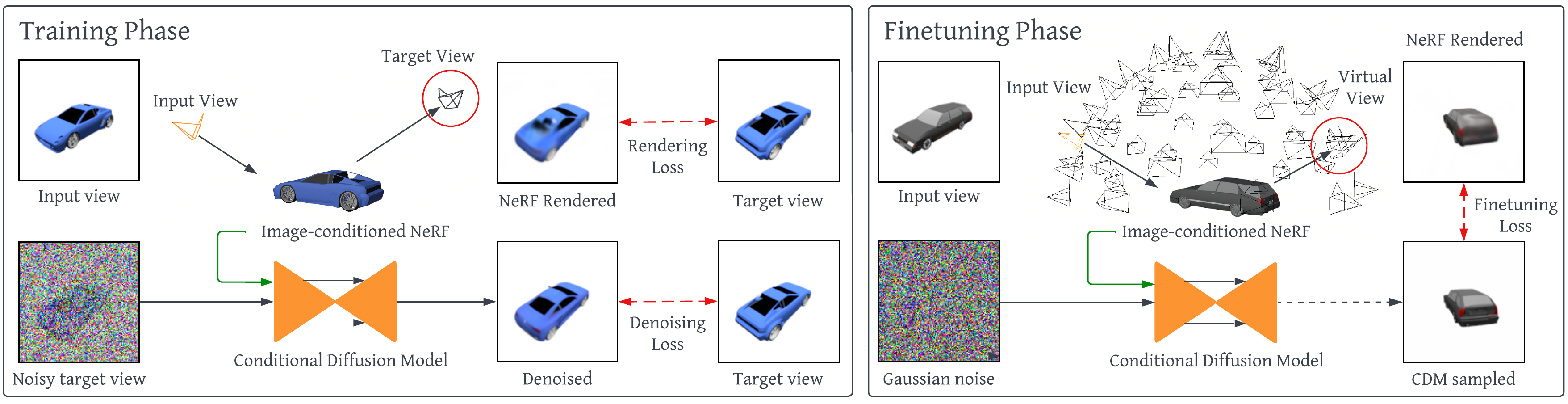}
\vspace{-10pt}
\caption{{\model} incorporates a training and finetuning pipeline. We first learn the single-image NeRF and 2D CDM, which are conditioned on the single-image NeRF renderings (left). We use the learned network parameters at test time to predict an initial NeRF representation for finetuning. The NeRF-guided denoised images from the frozen CDM then supervise the NeRF in turn (right).}
\vspace{-5pt}
\label{fig:example_pipeline}
\end{figure*}

In this paper, we propose {\model}, a \textit{training-finetuning} framework for synthesizing multi-view consistent and high-quality images given single-view input.
Concretely, at the training stage, we jointly train a camera-space triplane-based NeRF together with a 3D-aware conditional diffusion model (CDM) on a collection of scenes. 
We initialize the NeRF representation given the input image at the finetuning stage.
Then, we finetune the parameters from a set of virtual images predicted by the CDM conditioned on the NeRF-rendered outputs.
We found that a naive finetuning strategy of optimizing the NeRF parameters directly using the CDM outputs would lead to subpar renderings, as the CDM outputs tend to be multi-view inconsistent.
Therefore, we propose \textit{NeRF-guided distillation}, which updates the NeRF representation and guides the multi-view diffusion process in an alternating fashion. 
In this way, the uncertainty in single-image view synthesis can be resolved by filling in unseen information from CDM; in the meantime, NeRF can guide CDM for multi-view consistent diffusion. 
An illustration of the proposed pipeline is shown in~\cref{fig:example_pipeline}.

We evaluate our approach on three challenging benchmarks. Our results indicate that the proposed {\model} significantly outperforms all the existing baselines, achieving high-quality generation with multi-view consistency. See supplementary materials for video results.
We summarize the main contributions as follows:
\begin{itemize}
    \vspace{-5pt}
    \item We develop a novel framework -- {\model} which jointly learns an image-conditioned NeRF and a CDM, and at test time finetunes the learned NeRF using a multi-view consistent diffusion process (\cref{sec:training},\cref{sec:finetuning}).
    \vspace{-5pt}
    \item We introduce an efficient image-conditioned NeRF representation based on camera-aligned triplanes, which is the core component enabling efficient rendering and finetuning from the CDM (\cref{sec:image_encoder}). 
    \vspace{-5pt}
    \item We propose a 3D-aware CDM, which integrates volume rendering into 2D  diffusion models, facilitating generalization to novel views (\cref{sec:diffusion}).
    \vspace{-5pt}
\end{itemize}

\section{Related Work}

\subsection{Diffusion models for 3D generation}
Diffusion-based generative models~\cite{sohl2015deep,ho2020denoising,song2019generative} have recently become state-of-the-art on image synthesis~\cite{dhariwal2021diffusion} and have shown remarkable results in handling highly under-constrained tasks such as text-to-image~\cite{ramesh2022hierarchical,rombach2021highresolution,saharia2022photorealistic} and text-to-video generation~\cite{ho2022imagen}. 

More recently, diffusion models have also been proven effective in 3D generation tasks.
On the one hand, many methods propose to directly apply diffusion in 3D space, including point clouds~\cite{nichol2022point}, voxels~\cite{muller2022diffrf} and fields~\cite{anonymous2023diffusion}. 
Some of them learn diffusion in a latent space derived from 3D space~\cite{bautista2022gaudi,shue20223d}. However, a clear limitation of these approaches is that it requires 3D ground truth for learning the diffusion process, which is hard to acquire in the real environment.
On the other hand, several works propose to learn 3D representations from diffusion in 2D space. For example, 
\citet{li20223ddesigner} learns the geometry based on a 2-view diffusion model; \citet{anciukevicius2022renderdiffusion} designs an architecture that generates and renders an intermediate 3D representation for each diffusion step.
Concurrently related to our method, a series of work~\cite{poole2022dreamfusion,wang2022score,lin2022magic3d,zhou2022sparsefusion,deng2022nerdi} have proposed score distillation that learns 3D representation directly from pre-trained 2D diffusion models.

\subsection{Single-view Novel View Synthesis}
\paragraph{Methods beyond Neural Fields} 
Most initial attempts at single-view 3D reconstruction relied on ground truth training data to estimate the geometry of objects. These methods typically mapped an image to its depth or directly to a 3D shape~\cite{Eigen2014,Saxena2009,Fan2017,Tatarchenko2017,Tulsiani2017}. Some methods \cite{Kato2017,Yan2016,Loper:ECCV:2014} provide 3D reconstruction estimates without ground truth 3D supervision using differentiable renderers; however, these methods were limited to reconstructing only the geometry, not the appearance. Recently, other methods have allowed the rendering of novel views without regard for multiview consistency. For example, ENR~\cite{dupont2020equivariant} utilizes convolutions with a projection to decode 3D voxel features to RGB. Targeting more complex scenes, SynSin~\cite{Wiles2020SynSin:Image} makes use of a differentiable point cloud renderer and an inpainter to extrapolate to unseen areas. InfiniteNature~\cite{liu2021infinite} utilizes estimated depth to iteratively inpaint novel views along a camera trajectory. Other works, such as GeoFree~\cite{geofree} and Pixelsynth~\cite{pixelsynth}, utilize an autoregressive prior to inferring unseen areas of the scene. Finally, light-field-based methods like \citet{srt22} and \citet{suhail2022generalizable} condition transformers on features from input images and query rays to directly output colors or directly invert into a latent space~\cite{Sitzmann2021LightRendering}.

\vspace{-10pt}\paragraph{Methods based on Neural Fields} 
Many methods propose to use neural fields (e.g., neural radiance field~\citep[NeRF,][]{mildenhall2020nerf}) for this task. For example, SinNeRF~\cite{sinnerf} renders novel views near the input image using pseudo geometry. On the other hand, 
\citet{Sitzmann2019, sharf, codenerf, autorf}  incorporate global latent codes and apply test-time tuning to refine these codes. This process is very similar to inversion into the latent space of 3D NeRF-based GANs~\cite{gu2021stylenerf,eg3d,bautista2022gaudi, pix2nerf}. Note that it requires (estimated) camera poses at test time, which hinders high-quality results. Furthermore, the global bottleneck hinders capturing fine details, and due to the optimization of the input view, such methods also cannot handle occlusion appropriately. Finally, \textit{image-conditioned} methods (e.g., pixelNeRF~\cite{pixelnerf} and VisionNeRF~\cite{visionnerf}) directly utilize local image features to condition NeRF and are the most relevant to our method. Note that, like pixelNeRF, our method can perform view synthesis without pose annotation at test time.
We provide background on this type of method in the next section.

\section{Background} \label{sec:background}
\subsection{Image-conditioned NeRF}\label{sec:pixelnerf}
Neural radiance fields~\citep[NeRF,][]{mildenhall2020nerf} have been proven remarkably effective for novel view synthesis. 
NeRF defines an implicit function $f_\theta: (\vx, \vd) \rightarrow (\vc, \sigma)$ given a spatial location $\vx\in \mathbb{R}^3$ and ray direction $\vd\in \mathbb{S}^2$, where $\theta$ are the learnable parameters, $\vc$ and $\sigma$ are the color and density, respectively.
To render a posed image $\mI$, we march a camera ray through each pixel $\vr(t)=\vx_o + t\vd$ (where $\vx_o$ is the camera origin) and calculate its color via an 
approximation of the volume rendering integral: 
\begin{equation}
    \mI_\theta(\vr) = \int_{t_\text{n}}^{t_\text{f}}\omega(t)\cdot\vc_\theta(\vr(t), \vd)\diff t,
    \label{eq.nerf}
\end{equation}
where $\omega(t) = e^{-\int_{t_\text{n}}^t\sigma_\theta(\vr(s))\diff s}\sigma_\theta(\vr(t))$, $t_\text{n}$ and $t_\text{f}$ are the near and far bounds of the ray, respectively.
When multi-view images are available, $\theta$ can be easily optimized with the standard MSE loss: 
\begin{equation}
    \mathcal{L}^{\text{NeRF}}_\theta  = \mathbb{E}_{\mI\sim \textrm{data}, \vr\sim\mathcal{R}(\mI)}\|\mI_\theta(\vr) - \mI(\vr)\|^2_2,
    \label{eq.nerf_train}
\end{equation}
where $\mathcal{R}(\mI)$ is the set of rays that composes $\mI$.  
To capture high-frequency details, NeRF encodes $\vx$ and $\vd$ with sinusoidal positional functions $\xi_\textrm{pos}(\vx), \xi_\textrm{pos}(\vd)$. 
Recently, studies have shown that encoding functions with local structures like 
triplanes~\cite{eg3d} achieves significantly faster inference speed without quality loss.

\begin{figure*}[t]
\centering
   \includegraphics[width=\linewidth]{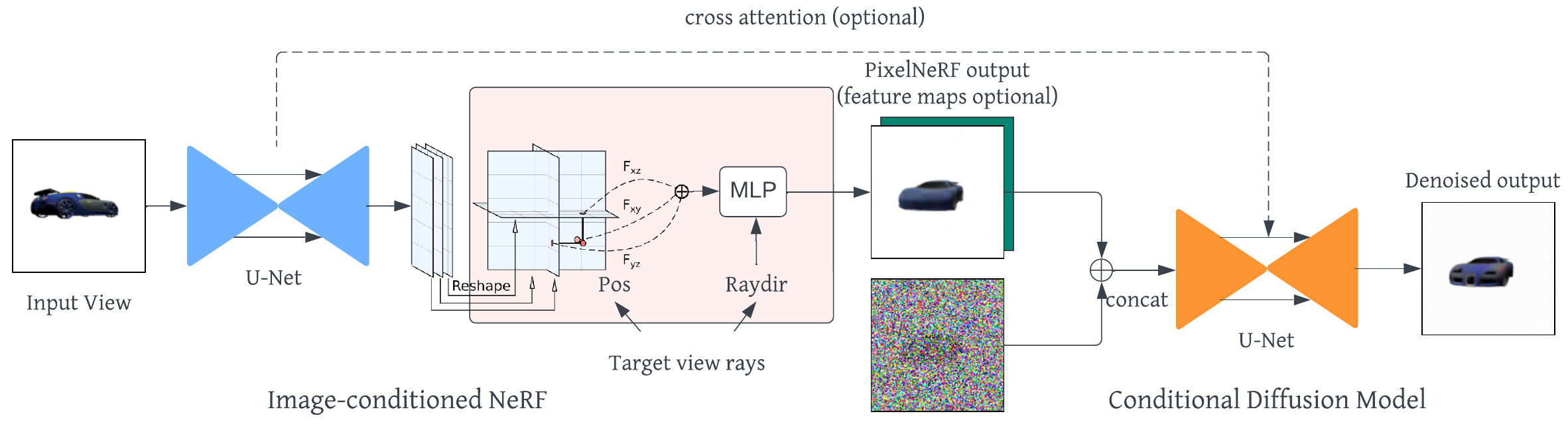}
   \vspace{-10pt}
   \caption{ Details of the architecture of the single-image NeRF for {\model}. Using a UNet, we first map an input image to a camera-aligned triplane-based NeRF representation. This triplane efficiently conditions volume rendering from a targeted view, resulting in an initial rendering. This rendering conditions the diffusion process so the CDM can consistently denoise at that target pose.}
   \vspace{-5pt}
\label{fig:pipeline}
\end{figure*}

The training of NeRF, i.e., the optimization of \Cref{eq.nerf_train}, requires tens or hundreds of images along with their camera parameters to provide sufficient multi-view constraints. 
However, in reality, such multi-view data is not easily accessible. Therefore, this work focuses on recovering neural radiance fields from a single image without knowing its absolute camera pose. 
As this problem is under-constrained, it requires 3D inductive biases learned from a large set of scenes similar to the target scene. 
Following this philosophy, pixel-aligned NeRFs~\cite{pixelnerf,visionnerf} encode 3D information with local 2D image features so that the learned representations can generalize to unseen scenes after being trained on a large number of scenes.
\vspace{-10pt}\paragraph{PixelNeRF}
Take PixelNeRF~\cite{pixelnerf} as an example. Given an input image $\mI^s$, PixelNeRF first extracts a feature volume $W=e_\psi(\mI^s)$ where $e_\psi$ is a learnable image encoder. Then, for any 3D point $\vx\in\mathbb{R}^3$ in the input \textit{camera space}, its corresponding image features are obtained by projecting $\vx$ onto the image plane as $\proj(\vx)\in[-1,1]^2$ with known intrinsic matrix, and then bilinearly interpolating the feature volume as $\xi_W(\vx) = W(\proj(\vx))$. 
The image features will be combined with the position $\vx$ and view direction $\vd$ to infer the color and density. Next, similar to NeRF, the color of a camera ray is calculated via volume rendering (Eq.~\ref{eq.nerf}). Such a model is trained over a collection of scenes, and for each scene, at least two views are needed to form the training pairs $(\mI^s, \mI)$ for reconstruction:
\begin{equation}
   \mathcal{L}_{\theta,\psi}^{\text{IC}}=\mathbb{E}_{(\mI^s, \mI)\sim \textrm{data}, \vr\sim \mathcal{R}(\mI)}\|\mI_{\theta, W}(\vr) - \mI(\vr)\|^2_2,
   \label{eq.pixelnerf}
\end{equation}
where  
$\mI_{\theta, W}$ is the volume rendered image. 
\vspace{-10pt}\paragraph{Challenges}
However, existing single-image NeRF approaches fail to produce high-fidelity rendering results, especially when 
severe occlusions exist.
This is because single-image view synthesis is an under-constrained problem, as the synthesized occluded regions can exhibit multiple possibilities.
Therefore, MSE loss (\cref{eq.pixelnerf}) forces single-image NeRF to regress to mean pixel values across all possible solutions, yielding inaccurate and blurry predictions. 

\subsection{Geometry-free View Synthesis}\label{sec:geofree}
To account for the uncertainty challenge, a distinct line of research explicitly models view prediction $p(\mI|\mI^s)$ with 2D generative models, like \citet{dupont2020equivariant,geofree,srt22} and more recently conditional diffusion models~\citep[3DiM,][]{watson2022novel}. 
Take 3DiM as an example.
It learns a conditional noise predictor $\veps_\phi$ that de-noises Gaussian-noised target images conditioning on the input view 
Moreover, the corresponding camera poses. Such a model can be optimized with a denoising loss:  
\begin{equation}
    \mathcal{L}^\text{DM}_\phi = \mathbb{E}_{(\mI^s, \mI)\sim \textrm{data}, \bm{\epsilon}, t}
    \|\veps_\phi\left(\mZ_t, \mI^s\right) - \veps\|^2_2
    \label{eq.ddpm}
\end{equation}
where 
$\mZ_t= \alpha_t \mI + \sigma_t \veps, \veps\sim \mathcal{N}(0, 1), \alpha_t^2+\sigma_t^2=1$ 
is the noised target for $\mI$.
As shown in~\citet{song2019generative}, the denoiser provides an approximation for the score function of the distribution $\veps_\phi(\mZ_t, \mI^s)\approx-\sigma_t\nabla_{\mZ_t}\log p_\phi(\mZ_t|\mI^s)$. 
At test time, the learned score $\veps_\phi$ is applied iteratively and refines noise images to synthesize novel views.

\vspace{-10pt}\paragraph{Challenges}
Geometry-free models typically suffer from the ``alignment problem'' where the input view conditioning and target views are not pixel-wise aligned, leading to inferior generalization when applying standard UNet-based diffusion models. 
\citet{watson2022novel} attempted to alleviate this issue by using cross-attention to gather information from the input view.
However, this requires models with large capacities, and even with this modification, it still needs more generalizability for complex scenes and out-of-distribution cameras.
Moreover, since denoising is conducted in 2D for each view independently rather than in 3D, the synthesized novel views of CDMs in the sampling stage tend to be multi-view inconsistent. 

\section{\model}
\label{section:method}

To achieve the best of both worlds, in this paper, we present a \textit{training-finetuning} two-stage approach, dubbed as \textit{{\model}}, to incorporate the power of diffusion models into image-conditioned NeRFs for single-image view synthesis. 
We illustrate the pipelines of the proposed two stages in \cref{fig:example_pipeline}.
In the following, we first introduce {\model}, which consists of an improved single-image NeRF based on local triplanes~(\cref{sec:image_encoder}) and a 3D-aware CDM built on top of the single-image NeRF outputs~(\cref{sec:diffusion}).  
An overview of the proposed model is presented in \cref{fig:pipeline}. 
These two components are optimized jointly on the same training set~(\cref{sec:training}).
At test time, we adopt a second-stage finetuning. Furthermore, to mitigate the inconsistency issue brought by CDM sampling, we present the \textit{NeRF-Guided Distillation} (NGD) algorithm to improve the finetuning performance~(\cref{sec:finetuning}).

\subsection{Single-image NeRF with Local Triplanes} \label{sec:image_encoder}
{\model} is built upon an efficient camera-aligned triplane extracted directly from an input image to condition the NeRF.
As mentioned in \cref{sec:pixelnerf}, most existing single-view models~\cite{pixelnerf,visionnerf} query the extracted features via image plane projection: $\proj: \mathbb{R}^3\rightarrow[-1,1]^2$.
One issue with this operation is that the depth information of a 3D point is not contained in its extracted features; that is, all points along the same camera ray project to the same location on the 2D image and thus have the same features. Therefore, to differentiate the points along the same camera ray, existing methods need to concatenate the image features with the positional encoding of the global spatial location $\xi_\textrm{pos}(\vx)$ as the representation of point $x$. However, this 3D representation is not efficient. Thus it needs a deep MLP network to fuse the image features with spatial information for inferring the color and density of $x$, which slows down the rendering process. 
Inspired by~\citet{eg3d}, 
we propose an efficient 3D representation that reshapes the image feature $W$ into a \emph{camera-aligned} triplane: $\{W_{xy}, W_{xz}, W_{yz}\}$\footnote{The $xy$ plane is aligned with the input image, while the $xz$ and $yz$ planes are orthogonal to the $xy$ plane and each other.}. Then, each 3D point receives a unique feature vector by bilinear interpolation within three planes:
\begin{equation}
    \xi_W(\vx) = W_{xy}(\tilde{\vx}_{xy}) + W_{xz}(\tilde{\vx}_{xz}) + W_{yz}(\tilde{\vx}_{yz}),
\end{equation}
where $\tilde{\vx} = \left[\proj(\vx), 
2 \cdot\frac{\vx_z - t_\text{n}}{t_\text{f}-t_\text{n}} - 1\right]\in [-1,1]^3$, and $t_\text{n}, t_\text{f}$ are the near and far bounds of the input camera (\cref{eq.nerf}). 
As this representation is expressive in the sense that it can allocate depth information in the $xz, yz$ planes, no additional positional encoding $\xi_\textrm{pos}(\vx)$ is needed to augment the representation, and 
the deep MLP network can be replaced with a shallow MLP network. 
This not only makes high-resolution image rendering efficient (\cref{sec:diffusion}) but also enables fast NeRF finetuning, which will be elaborated in \cref{sec:finetuning}.
Furthermore, modeling triplanes in the camera space of the input image has the following benefits: (1) The triplane can naturally preserve the local image features, same as  pixelNeRF~\cite{pixelnerf}; (2) we do not need to assume a global coordinate system, and global camera poses, which is different from existing triplane-based methods~\cite{eg3d,tensorf,bautista2022gaudi}. 

Note that, for the image encoder, we adopt a UNet architecture~\cite{Ronneberger2015,nichol2021improved} rather than a pre-trained ResNet~\cite{He2016b} used in \citet{pixelnerf}. 
Thanks to the U-connection and self-attention blocks, 
the output layer feature $W=e_\psi(I)$ 
contains both the local and global information that is essential for predicting occluded views, which works similarly to the feature extractors in \citet{visionnerf}. See \cref{fig:pipeline} for details.

\subsection{3D-aware Conditional Diffusion Models}\label{sec:diffusion}
While {\pixelnerf} produces multi-view consistent images, the outputs tend to be blurry due to the uncertainty issue 
(\cref{sec:pixelnerf}). 
To address the uncertainty issue, we model a 3D-aware CDM as the second part of {\model}, which resolves uncertainty through a generative process.
Specifically, the CDM is learned to iteratively refine the rendering of {\pixelnerf} to match the target views.

Compared to existing geometry-free methods~\cite{watson2022novel}, 
we avoid the ``alignment" problem by applying {\pixelnerf} to render the target-view images as the conditioning to CDM rather than using the input-view image as conditioning.
As shown in \cref{fig:pipeline}, we adopt the standard conditional UNet architecture~\cite{nichol2021improved} where the noisy image is concatenated with the rendered image.
Similar to \citet{watson2022novel}, we can also employ cross-attention blocks between the CDM UNet and the encoder UNet
to strengthen the conditioning. Note that the efficiency of the triplane rendering (see~\cref{table:speedresults}) allows the NeRF to be trained in tandem with the CDM, which would take far too long otherwise.

\setlength{\textfloatsep}{5pt}
\begin{algorithm}[t]
\small
\DontPrintSemicolon
  \KwInput{NeRF (MLP $f_\theta$, triplanes $W$), CDM $\veps_\phi$, input $\mI^s$, $\gamma$, $N, B$}
  \textbf{Initialize} $I^\pi=I_{\theta, W}^\pi, \veps^\pi=\veps, \pi\in \Pi, \veps\sim \mathcal{N}(0,1)$\\
  \For{$t=t_{\max} \ldots t_{\min}$} {
    \For{$\pi \in \Pi$}
     { 
        {$\mZ^\pi = \alpha_t \mI^\pi + \sigma_t \veps^\pi$}; \\
        {$\veps^\pi =\veps_\phi(\mZ^\pi,\mI^s) + \gamma\sigma_t/\alpha_t\cdot(\mI^\pi - \mI^\pi_{\theta, W})$} \\
        {$\mI^\pi = (\mZ^\pi - \sigma_t\veps^\pi) / \alpha_t$}
    }
    \For{$n=1\ldots N$}{
        \For{$b=1\ldots B$}{
            {Sample a view $\pi\sim \Pi$ and sample a ray $\vr$ from $\pi$;}}
        {Update $\theta, W$ with $\nabla_{\theta,W}\frac{1}{B}\sum_{\pi,\vr}\|\mI^\pi_{\theta, W}(\vr) - \mI^\pi(\vr) \|^2_2$}
    }
}
\Return{$\theta, W$}
\caption{\bf Finetuning with NeRF-guided distillation.\label{alg.sampling}}
\end{algorithm}

\subsection{Training Phase}\label{sec:training}
Given a collection of scenes where each scene has at least two views, we train end-to-end by combining \cref{eq.pixelnerf,eq.ddpm}: $\mathcal{L}^{\text{Train}}_{\theta,\psi,\phi} = \lambda_{\text{IC}}\mathcal{L}_{\theta,\psi}^{\text{IC}} +  \lambda_{\text{DM}}\mathcal{L}^\text{DM}_\phi$ to optimize the {\pixelnerf} and CDM jointly.
Note that the training of {\pixelnerf} receives supervision from both the photometric error $\mathcal{L}_{\theta,\psi}^{\text{IC}}$ and  the CDM denoising loss $\mathcal{L}^\text{DM}_\phi$. See Figure~\ref{fig:example_pipeline} (left) for details.

\setlength\tabcolsep{2.1pt}
\begin{table*}[tbh]
    \small
    \centering
    \begin{tabular}{l  cccc cccc cccc}
         \toprule
         &  \multicolumn{4}{c}{\bf ShapeNet Cars} &  \multicolumn{4}{c}{\bf ShapeNet Chairs} &  \multicolumn{4}{c}{\bf Amazon-Berkeley Objects} 
         \\
         & PSNR$\uparrow$ & SSIM$\uparrow$ & LPIPS$\downarrow$ & FID$\downarrow$  & PSNR$\uparrow$ & SSIM$\uparrow$ & LPIPS$\downarrow$  & FID$\downarrow$   & PSNR$\uparrow$ & SSIM$\uparrow$ & LPIPS$\downarrow$  & FID$\downarrow$\\
         \midrule
         LFN~\cite{Sitzmann2021LightRendering}$^*$ & 22.42 & 0.89 & -- & -- & 22.26 & 0.90 & -- & -- & -- & -- & -- & -- \\
         3DiM~\cite{watson2022novel}$^*$  & 21.01 & 0.57 & --    & {\bf 8.99}  & 17.05 & 0.53 & --    & 6.57 & -- & -- & -- & --  \\
         \midrule
         SRN~\cite{Sitzmann2019}          & 22.25 & 0.88 & 0.129 & 41.21 & 22.89 & 0.89 & 0.104 & 26.51 & -- & -- & -- & -- \\
         PixelNeRF~\cite{pixelnerf}       & 23.17 & 0.89 & 0.146 & 59.24 & 23.72 & 0.90 & 0.128 & 38.49 & -- & -- & -- & -- \\
         CodeNeRF~\cite{codenerf}         & 22.73 & 0.89 & 0.128 & --    & 23.39 & 0.87 & 0.166 & -- & -- & -- & -- & -- \\
         FE-NVS~\cite{guo2022fast}        & 22.83 & 0.91 & 0.099 & --    & 23.21 & 0.92 & 0.077 & -- & -- & -- & -- & --\\
         VisionNeRF~\cite{visionnerf}     & 22.88 & 0.90 & 0.084 & 21.31 & 24.48 & 0.92 & 0.077 & 10.05 & 28.61 & 0.93 & 0.095 & 33.38 \\
         \midrule
         
         {\model}-B (Ours)         & 23.51 & {\bf 0.92} & {0.082} & {18.09} & 24.79 & 0.94 & {\bf 0.056} & 5.65 & 32.81 & 0.96 & 0.057 & 7.77 \\
         \; \; \;  w/o NGD    & {23.81} & {\bf 0.92} & 0.093 & 42.37 & 24.77 & 0.93 & 0.068 & 15.72 & 32.07 & 0.95 & 0.063 & 18.01 \\

          {\model}-L (Ours)         & 23.76 & {\bf 0.92} & {\bf 0.076} & {15.49} & {\bf 24.95} & {\bf 0.94} & {\bf 0.056} & {\bf 5.34} & {\bf 32.84} & {\bf 0.97} & {\bf 0.042} & {\bf 6.31}\\
         \; \; \; w/o NGD     & {\bf 23.95} & {\bf 0.92} & 0.092 & 43.26 & 24.80 & 0.93 & 0.070 & 15.50 & 32.00 & 0.96 & 0.061 & 17.73\\

         \bottomrule
    \end{tabular}
    \caption{Comparisons on ShapeNet Cars \& Chairs and ABO datasets across baselines. $^*$ indicates geometry-free model. The results of the baselines except VisionNeRF~\cite{visionnerf} are copied from the official papers. {\bf --} denotes the results are unavailable. }
    \label{tab:results}
\end{table*}
\begin{table}[t]
\centering
\small
\begin{tabular}{lcccc}
\toprule
Method & PSNR$\uparrow$ & SSIM$\uparrow$ & LPIPS$\downarrow$ & FID$\downarrow$\\
\midrule
VisionNeRF~\cite{visionnerf} & {\bf 35.94} & {\bf 0.97} & 0.065 & 11.18 \\
{\model}-B (Ours) & 34.81 & {\bf 0.97} & {\bf 0.040} & {\bf 6.76} \\
\bottomrule
\end{tabular}
\caption{Quantative results on Clevr3D} 
\label{table:clevrresults}  
\end{table}
\begin{figure*}[t]
    \centering
    \includegraphics[width=0.95\linewidth]{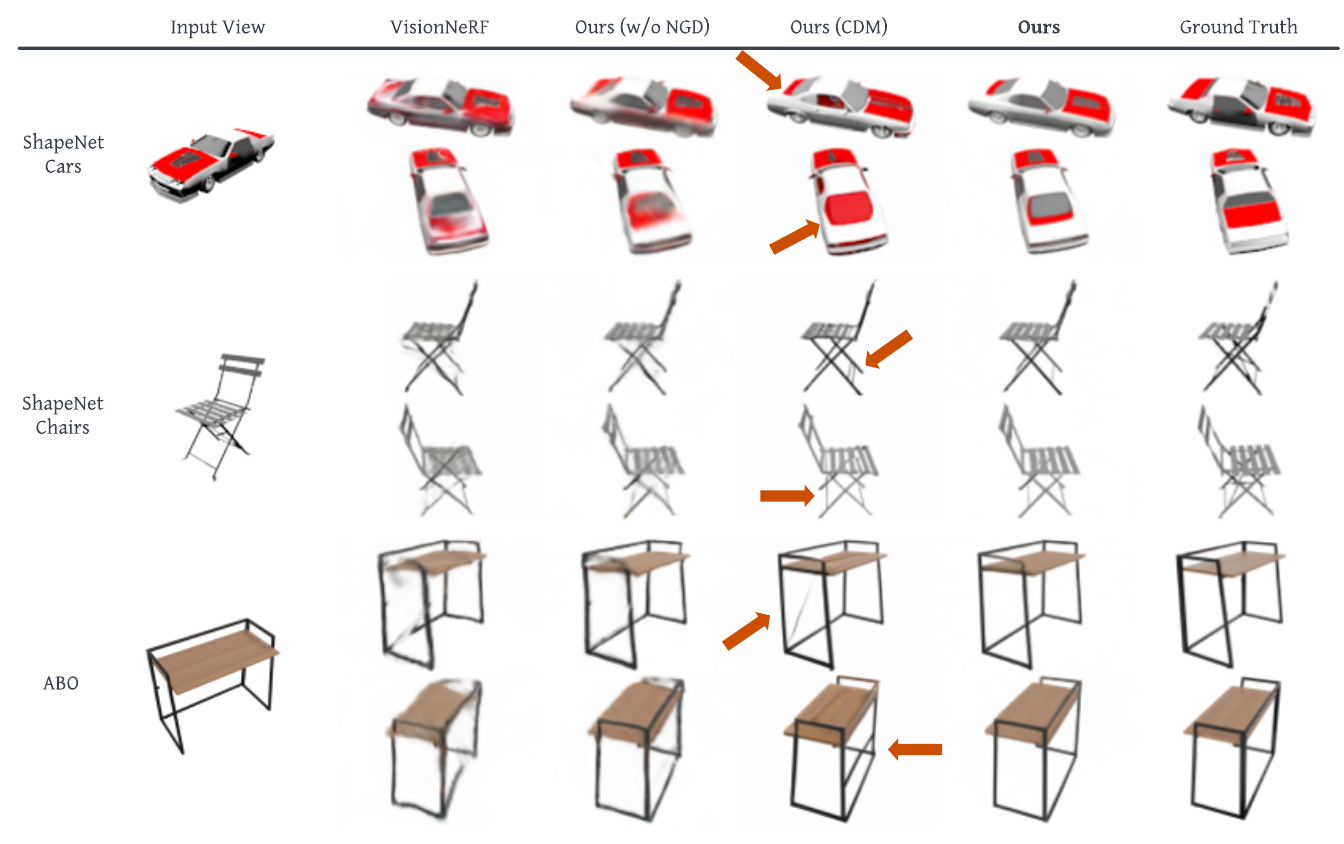}
\caption{A qualitative comparison of our approach versus baselines in single-image view synthesis on multiple datasets. 
Compared to 3D methods like VisionNeRF~\cite{visionnerf} and Ours(w/o NGD), our proposed {\model} synthesizes significantly sharper results behind occlusions. Compared to Ours (CDM), our full model showcases its built-in multi-view consistency. The red arrows display the CDM's inability to synthesize consistently across views. }
    \label{fig:comparison}
\end{figure*}
\begin{figure*}[t]
    \centering
    \includegraphics[width=0.95\linewidth]{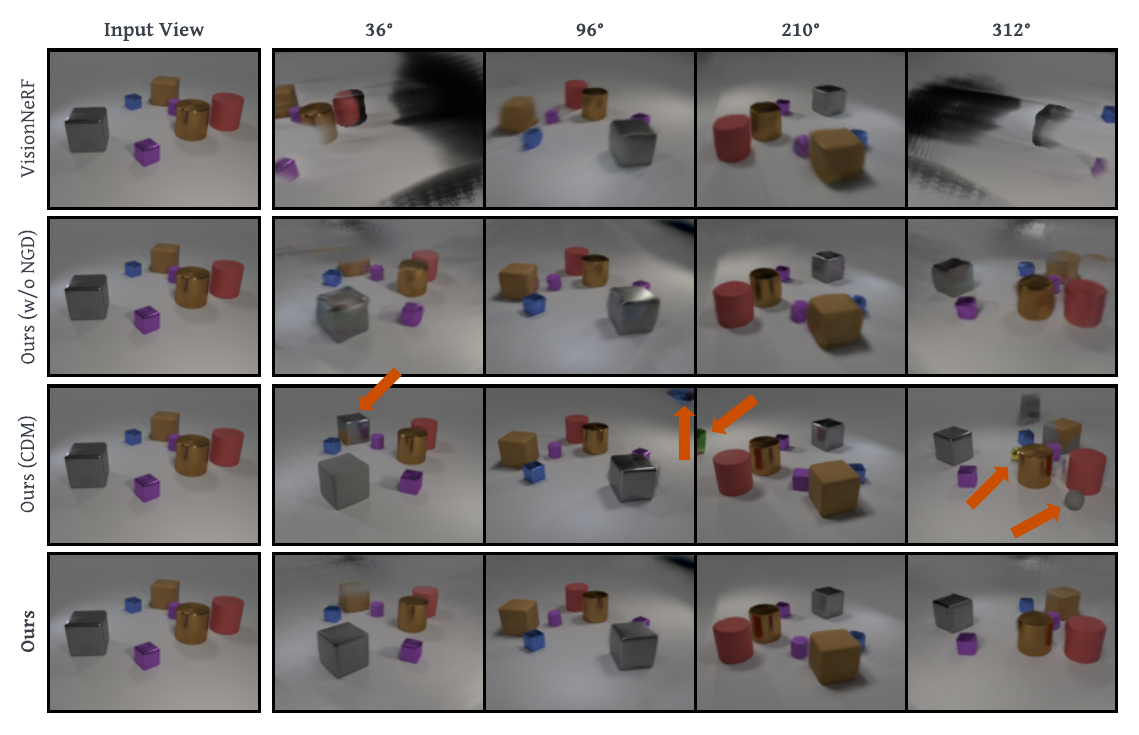}
\caption{A qualitative comparison on Clevr3d~\cite{obsurf} which consists of images from cameras rotated 120 degrees about the \textit{z}-axis. We showcase generalization to OOD cameras in this figure. As can be seen, VisionNeRF gets a degenerate result, while {\model} provides sharper renderings with fewer artifacts.}
    \label{fig:comparison_clevr}
\end{figure*}

\subsection{Fine-tuning Phase}\label{sec:finetuning}
While the 3D-aware CDM resolves the uncertainty issue in the single-image NeRF and thus makes the synthesized images sharper, it compromises multi-view consistency, as the 2D diffusion process is independently applied to each novel view. To synthesize multi-view consistent and high-quality results, we propose a novel finetuning strategy at test time to \textit{distill} the CDM's knowledge.

As shown in \cref{fig:example_pipeline} (right), given an input view $\mI^s$ of an unseen scene, we generate a set of ``virtual views" with the trained single-image NeRF and 3D-aware CDM. Then we finetune the triplane parameters $W=e_\psi(\mI^s)$ and MLP parameters $\theta$ (pink box in Fig.~\ref{fig:pipeline}) with the generated virtual views.  
Here, we treat $W$ as learnable parameters.
It performs best when virtual views cover the region of interest. 
\vspace{-10pt}\paragraph{NeRF Guided Distillation} A naive  optimization strategy is the same as that in ~\citet{mildenhall2020nerf} (\cref{eq.nerf_train}), i.e., replacing the targets with the ``virtual views" sampled from the CDM.
We found that this naive optimization typically leads to noisy results with severe floating artifacts, as the inconsistent CDM predictions cause conflicts in optimizing the NeRF model. 
Instead, we propose \textit{NeRF Guided Distillation (NGD)} that alternates between NeRF distillation and diffusion sampling. 
Inspired by classifier guidance~\cite{dhariwal2021diffusion}, we incorporate 3D consistency into multi-view diffusion by considering the joint distribution for each virtual view $\mI$: 
\begin{equation}
    \begin{split}
    p_\phi(\mZ_t, \mI_{\theta,W}|\mI^s) &= p_\phi(\mZ_t|\mI^s)\cdot p(\mI_{\theta,W} | \mZ_t,\mI^s)\\&\propto p_\phi(\mZ_t|\mI^s)\cdot e^{-\frac{\gamma}{2}\|\mI_t - \mI_{\theta,W}\|^2_2},
    \end{split}
    \label{eq.modified_prob}
\end{equation}
where $\mI_t = (\mZ_t -\sigma_t\veps_\phi(\mZ_t, \mI^s))/\alpha_t$ is the predicted target image at the intermediate timestep. The second term introduces multi-view constraints from a given NeRF. 
Therefore, the goal is to find NeRF parameters ($\theta, W$) that maximize \cref{eq.modified_prob} while sampling the most likely virtual views ($\mZ_t$) from the joint distribution. 
In practice, we adopt an iterative-based updating rule at each diffusion step $t$.
For generating virtual views with the CDM, we follow the modified diffusion score derived from \cref{eq.modified_prob}: 
\begin{equation}
    \tilde{\veps}_\phi(\mZ_t, \mI^s) = \veps_\phi(\mZ_t, \mI^s) + \gamma\frac{\sigma_t}{\alpha_t}(\mI_t-\mI_{\theta, W}),
    \label{eq.eps_guide}
\end{equation}
where $\tilde{\veps}_\phi$ will be used in regular DDIM sampling~\cite{song2020denoising} \footnote{We consider ${\partial \mI_t}/{\partial \mZ_t}\approx 1/\alpha_t$ to avoid backpropagation through the UNet, similar to DreamFusion~\cite{poole2022dreamfusion}. }.
Note that for $\gamma={\alpha_t^2}/{\sigma_t^2}$ (SNR), following the modified score~\cref{eq.eps_guide} is equivalent to replacing the denoised images with the NeRF rendering. 
For distilling NeRF, we directly maximize the log-likelihood of this joint distribution w.r.t. the NeRF parameters, which is equivalent to minimizing the MSE loss between the denoised images $\mI_t$ and the NeRF renderings $\mI_{\theta,W}$ across all virtual views:
\begin{equation}
    \mathcal{L}^{\text{FT}}_{\theta, W}=\mathbb{E}_{\pi\sim \Pi, \vr\sim \mathcal{R}(\mI^\pi_t)} \|\mI^\pi_{\theta, W}(\vr) - \mI_t^\pi(\vr) \|^2_2,
    \label{eq.finetune}
\end{equation}
where $\Pi$ is a prior distribution on the relative camera poses to the input and $\mI_t^\pi,\mI^\pi_{\theta, W}$ are the corresponding images
at the relative camera $\pi$.
Note that to reduce computation, we sample the rays $\vr$ with batch size $B$ from all views, supervise only the corresponding pixels, and finetune for $N$ steps. 
The algorithm details are shown in \cref{alg.sampling}.
\vspace{-10pt}\paragraph{Relationship to SDS}
Our method shares similarities with the recently proposed \textit{score distillation sampling}~\citep[SDS,][]{poole2022dreamfusion}. 
Although SDS also \textit{distills} the diffusion models into 3D, there is a fundamental difference.
In SDS, a random-scaled noise is injected into NeRF's output from a random angle.
The noised image is then denoised by a 2D diffusion model to provide supervision. 
In contrast, our method initializes a set of virtual views and uses NeRF to guide the diffusion process of each view (and alternatingly refines NeRF based on this diffusion). As a result, our pipeline completes the full diffusion trajectory for every view, following a naturally decreasing noise schedule. 
In \cref{section:comparisondetails}, we show additional comparisons and potential reasons SDS is practically worse than our method.

\section{Experiments} \label{sec:results}

\subsection{Experimental Settings}
\begin{table}[t]
\centering
\small
\begin{tabular}{lcc}
\toprule
Method & Image encoding & Rendering\\
\midrule
PixelNeRF~\cite{pixelnerf}  & 0.007s & 1.639s \\
VisionNeRF~\cite{visionnerf} & 0.015s &  0.678s \\
\midrule
{\model}-B & 0.024s & 0.018s \\
{\model}-L & 0.031s & 0.018s \\
\bottomrule
\end{tabular}
\caption{Comparison of encoding and rendering speed on ShapeNet Cars dataset between models.} 
\label{table:speedresults}  
\end{table}
\paragraph{Datasets}
We evaluate {\model} on three benchmarks -- SRN-ShapeNet~\cite{Sitzmann2019}, Amazon-Berkeley Objects~\citep[ABO,][]{collins2022abo} and Clevr3D~\cite{obsurf} -- 
for testing novel view synthesis under single-category, category-agnostic, and multi-object settings, respectively. 
SRN-ShapeNet includes two categories: \textit{Cars} and \textit{Chairs}. Dataset details are given in \cref{subsection:datasets}.
\vspace{-10pt}\paragraph{Baselines}
We choose the pixel-aligned method VisionNeRF~\cite{visionnerf} as the main baseline for comparison considering its state-of-the-art performance in single-image view synthesis. 
We additionally evaluate our proposed {\pixelnerf} without the fine-tuning stage (denoted as ``Ours (w/o NGD)"), 
Furthermore, show qualitative results from the CDM prediction without NeRF guidance  (denoted as ``Ours (CDM)").
Besides, we include publicly-available results for other methods such as SRNs~\cite{Sitzmann2019}, CodeNeRF~\cite{codenerf}, FE-NVS~\cite{guo2022fast}, and geometry-free approaches  LFN~\cite{Sitzmann2021LightRendering} and 3DiM~\cite{watson2022novel}.
\vspace{-10pt}\paragraph{Evaluation Metrics}
We evaluate our model and the baselines by comparing the generated images and target views given a single image, and the relative target camera poses as input. We report four standard metrics: PSNR, SSIM, LPIPS~\cite{Zhang2018f}, and FID~\cite{fid}. PSNR measures the mean-squared error per pixel, while SSIM measures the structural similarity; LPIPS is a deep metric that reflects the perceptual similarity between images. Finally, FID measures the similarity between the distribution of the rendered and ground truth images of all test scenes. Note that generative frameworks--due to their multimodal nature--generally perform poorly with respect to PSNR, which prioritizes proximity to the mean pixel values.

\begin{table}[t]
\centering
\small
\begin{tabular}{lcccc}
\toprule
Method & PSNR$\uparrow$ & SSIM$\uparrow$ & LPIPS$\downarrow$ & FID$\downarrow$\\
\midrule
No Fine-tuning &{23.81} & 0.915 & 0.093 & 42.37\\
\midrule
Fine-tuning \\ 
\; \; Direct distillation& 23.46 & 0.911 & 0.105 & 35.88 \\
\; \; w. SDS~\cite{poole2022dreamfusion} & 20.32 & 0.882 & 0.106 & 28.06 \\
\; \; w. NGD (Ours) & {\bf 23.51} & {\bf 0.917} & {\bf 0.082} & {\bf 18.09}\\
\bottomrule
\end{tabular}
\caption{Ablation on fine-tuning strategy on ShapeNet-Cars.} 
\label{table:finetuneresults}  
\end{table}
\begin{table}[t]
\centering
\small
\begin{tabular}{l|ccccc|c}
\toprule
\# virtual views  & 5 & 10 & 25 & 50 & 100  & w/o NGD\\
\midrule
PSNR$\uparrow$ &  22.86 & 23.16 & 23.34 & 23.51 & {\bf 23.55} & 23.81 \\
SSIM$\uparrow$ &  0.901 & 0.913 & 0.915 & {\bf 0.917} & 0.916 & 0.915\\
LPIPS$\downarrow$ &  0.095 & 0.085 & {\bf 0.083} & {\bf 0.083} & 0.087 & 0.093\\
FID$\downarrow$ & 27.41 & {\bf 16.04} & 17.14 & 18.09 & 19.13 & 42.37\\
\bottomrule
\end{tabular}
\caption{Comparison on the number of virtual views used for fine-tuning on ShapeNet Cars.} 
\label{table:virtualresults}  
\end{table}
\subsection{Main Results}
We show results with two variant sizes ({\model}-B:$\sim400$M parameters, {\model}-L:$\sim1$B parameters). Details of the implementation specifics are given in \cref{section:implement}.
\vspace{-10pt}
\paragraph{Quantitative evaluation}
\cref{tab:results,table:clevrresults} show the quantitative comparisons of our proposed models to the SoTA geometry-free and single-view NeRF methods on all three datasets.
The quantitative scores of the baselines are copied from the official papers if available. 
Our proposed {\model} (with and without NGD finetuning) significantly outperform all baselines in PSNR and SSIM, displaying the models' ability to synthesize accurate pixel-level details with the local triplane representations. Additionally, in LPIPS, our proposed {\model} is better than all previous approaches indicating its ability to create perceptually correct completions behind occlusions. Finally, about FID, our method outperforms all single-view NeRF methods, only having worse scores than 3DiM on ShapeNet-Cars as it is purely 2D. 
Note that, as mentioned in the original paper~\cite{watson2022novel}, 3DiM cannot generalize well to the out-of-the-distribution testing cameras of ShapetNet-Chairs, thus performing poorly. 
In contrast, with the 3D-aware CDM, our approach can easily handle unseen viewpoints.
In addition, the proposed NGD finetuning, while slightly hurting PSNR in some cases, significantly improves the sharpness of the results, thus resulting in better FID and LPIPS scores. Besides, scaling the model size up further yields higher perceptual quality.
\vspace{-10pt}\paragraph{Qualitative evaluation}
\cref{fig:comparison} displays the qualitative comparison of our approach to the main baseline, VisionNeRF~\cite{visionnerf}, and two ablated models. Our method produces much more detailed results than the ablated model {\pixelnerf} and VisionNeRF on ShapeNet and ABO. Due to their reliance on projected image features, these methods cannot handle uncertainty behind occlusion and thus regress mean pixel values, resulting in blurry renderings. The CDM results are worse aligned and inconsistent across views, 
as demonstrated by the red arrows in \cref{fig:comparison}.
\cref{fig:comparison_clevr} shows additional qualitative results on Clevr3D. 
Our method again shows consistent and high-quality renderings. At the same time, VisionNeRF overfits the camera distribution and fails to synthesize viewpoints close to the input (see \cref{section:implement} for more details). The CDM results are again inconsistent with objects appearing and disappearing.
\textbf{Please refer to the supplementary materials for uncurated and extensive video results} showing the multiview consistency and high fidelity of our method.

\subsection{Ablation Studies}\label{subsection:ablations}
We provide ablations on the Shapenet-Cars dataset to validate our model's key design choices, making our ablation results directly comparable to the ShapeNet Cars results in \cref{fig:comparison}. In \cref{table:finetuneresults}, we compare our model without any CDM-based finetuning and various CDM-based finetuning strategies. As seen in the results, finetuning with a CDM will improve the unconditional FID. However, only our NGD sampling will yield the state-of-the-art conditional SSIM and LPIPS. For details of the sampling baselines Direct Distillation and SDS compared to our NGD, please see \cref{section:comparisondetails}. \cref{fig:ablation_sample} also provides a qualitative comparison. Next, in \cref{table:virtualresults}, we also provide ablations on the number of virtual views for finetuning. With too few (e.g.
5) virtual views, the NeRF overfits the denoised images resulting in subpar renderings. We find that 50 virtual views provide a good tradeoff between efficiency and performance.


\section{Discussion}
\paragraph{Limitations}
Our proposed method has two main limitations. Firstly, we require at least two views of a scene at training time. Secondly, our finetuning process is expensive in time, limiting application in real-time domains. Future work may address these issues.
\vspace{-10pt}\paragraph{Future work}
 For future research, it is also possible to investigate our proposed NGD to improve the fidelity of text-to-3D pipelines \cite{dreamfields,poole2022dreamfusion, lin2022magic3d}. Additionally, more complex datasets such as the Waymo Open Dataset~\cite{waymodataset} may be explored, leaving the challenging task of occlusion-handling to large-scale pretrained 2D diffusion models as we do in this paper. 
Also, incorporating our finetuning strategy in the context of 3D GANs~\cite{gu2021stylenerf, eg3d} may improve inversion performance and 3D editing capabilities as well.
Finally, it may also be interesting to figure out how to properly incorporate priors such as symmetry.
\section{Conclusion}
We introduced {\model}, a generative framework for single-image view synthesis which distills a 3D-aware CDM to a triplane-based image-conditioned NeRF. 
We further introduced NeRF-guided distillation to sample multiple views from the CDM while simultaneously improving the NeRF renderings. Our method achieved the state-of-the-art results on multiple challenging benchmarks.

\bibliography{egbib}
\bibliographystyle{icml2022}

\newpage
\appendix
\newpage
\section*{\huge Appendix}
\section{Datasets} \label{subsection:datasets}
We validate our {\model} algorithm across three datasets. The details of each are given in the following subsections.

\subsection{ShapeNet}  
ShapeNet-Cars and -Chairs~\cite{Sitzmann2019} are standard for few-shot view synthesis benchmarking. We use the data hosted by pixelNeRF~\cite{pixelnerf}, which can be downloaded from GitHub (\url{https://github.com/sxyu/pixel-nerf}). 
The chairs dataset consists of 6591 scenes, and the cars dataset has 3514 scenes, both with a predefined train/val/test split. Each training scene contains $50$ posed images taken from random points on a sphere. Each testing scene contains $250$ posed images taken on an Archimedean spiral along the sphere. All scenes share intrinsic, and images are rendered at a resolution of $(128,128)$. At testing, we choose one pose as input (index 64) and keep this camera input constant for all scenes.

\subsection{Amazon Berkeley Objects (ABO)} 
We also consider the ABO dataset~\cite{collins2022abo} from \url{https://amazon-berkeley-objects.s3.amazonaws.com/index.html} under the title "ABO 3D Renderings." We randomly sampled a custom split. The dataset thus consists of 6743 training scenes, 396 validation scenes, and 794 testing scenes. Each scene consists of 30 images of an object rendered onto a white background in a physically-based manner. The objects are drawn from 64 different object categories, providing an extensive evaluation of the generalization capabilities of various models. The images have resolution $(256,256)$, and we crop and adjust the intrinsics so that the models are fed with images of size $(128,128)$. The cameras are not uniformly distributed, but all point at the object. For testing, we use an input camera index of 0.

\subsection{Clevr3D}
We consider the Clevr3D dataset
provided in \cite{obsurf} for multi-object/scene level learning, which can be downloaded from the github~\url{https://github.com/stelzner/obsurf}. We define a custom split in which there are
$70000$ training scenes and $1000$ held-out testing scenes. Each scene
consists of $3$ posed images at a resolution of $(120,160)$ with the camera pointing at the origin. These
images are rendered at $120$ degree rotations about the z-axis with varying distances from the origin. We
select one input image (index 0) at testing and render the other two views.

\section{Implementation Details} \label{section:implement}
\subsection{Architecture and Hyperaparameters}
For all datasets, we learn {\model} based on the U-Net architecture adopted from ADM~\cite{dhariwal2021diffusion} with two sets of configurations (-B: base $\sim400$M parameters, -L: large $\sim1$B parameters). More specifically, we set the model dimension $d=192$ with $2$ residual blocks per resolution for the base architecture and $d=256$ with $3$ residual blocks per resolution for the large architecture. All other hyperparameters follow the default setting as ADM.

Note that the image encoder retains the same architecture and hyperparameters as the CDM outlined above. Similar to \cite{watson2022novel}, we incorporate a cross-attention module between the CDM and the image encoder after every attention block to strengthen the conditioning.
The last layer output of the image encoder is reshaped to a triplane.
As a result, the triplane has the same spatial resolution as the input image, and we set the feature dimension of the triplane as $48$.
We implement the NeRF module (pink box in Fig.~\ref{fig:pipeline}) 
using a 2-layer MLP with a hidden size of $64$. 
For NeRF rendering, we follow \citet{visionnerf} and uniformly sample $64$ points along each ray, with $64$ additional points by importance sampling.
As mentioned in \cref{sec:diffusion}, we directly concat the NeRF rendered image with the noised input and send it to the CDM for denoising.

\subsection{Training phase}

The CDM is trained with cosine noise schedule $\alpha_t=\cos(0.5\pi t)$ based on velocity prediction~\cite{salimans2022progressive}.
We set $\lambda_\text{IC}=\lambda_\text{DM} = 1$, which means that we add the two losses of the two modules without re-weighting.
All models are trained using AdamW~\cite{loshchilov2017decoupled} with a learning rate of $2e{-5}$ and an EMA decaying rate of $0.9999$. 
We train all models with a batch size of $32$ images for 500K iterations on $8$ A100 GPUs.
Training takes $3-4$ days to finish for base models.

Note that for the Clevr3D dataset~\cite{obsurf}, we noticed that the models tended to overfit to the input view easily, creating a plane of density orthogonal to the camera axis and thus clearly degenerate geometry. To ameliorate this, we trained input view reconstruction with slightly noisy camera locations (variance $0.3$). We found that this fixed the issue for our method, but it still failed for VisionNeRF~\cite{visionnerf}, even after increasing the noise.

\subsection{Finetuning phase}
When finetuning with NGD, we define $K$ virtual views relative to the input image by sampling near the test trajectory. By default, we set $K=50$ for shapeNet and Clevr3D and $30$ for ABO.
See \cref{pi} for specific $K$ per dataset and how to obtain these poses. 
For the multiview diffusion process, we run $64$ DDIM~\cite{song2020denoising} steps with the CDM for each view, respectively. 
At every diffusion step, we update the NeRF parameters $N=64$ steps, with a batch size of $B=4096$ rays. We use Adam optimizer~\cite{Kingma2015a} set the learning rates for NeRF MLPs $1e-4$ and the triplane features $5e-2$, respectively. Our empirical results indicate that a large learning rate on a triplane can boost the finetuning efficiency.

\section{Prior Relative Camera Distribution $\Pi$} \label{pi}
In order to approximate the expectation in \cref{eq.finetune}, we require a sampling of $\Pi$, i.e., a sampling of $K$ 'important' or 'relevant' cameras, which adequately capture the region of interest. For each of the three datasets, we rely on the relative (to the input) camera poses of the testing set for this. The cameras are very different between datasets, requiring a slightly different procedure.

\paragraph{ShapeNet}
Because the Shapenet testing trajectory is an Archimedean spiral around the object consisting of 251 views, we simply uniformly sample every 5th camera yielding 50 cameras total from which we can approximate $\Pi$, thus yielding $K=50$ cameras to approximate \cref{eq.finetune}.

\paragraph{Amazon Berkeley Objects (ABO)} 
Because each scene contains only 30 cameras, we use all the relative poses of the testing set ($K=30$) to approximate \cref{eq.finetune}.

\paragraph{Clevr3D} As Clevr3D contains only three cameras per scene, creating a good sample of $\Pi$ is slightly more difficult. As we know, all of the Clevr3D cameras are pointing at the origin; we can calculate the relative position of the world origin in camera coordinates by intersecting two of the optical axes of the relative cameras. This will serve as the look-at-point for our virtual cameras. Note that the Clevr cameras are additionally all of a similar height relative to the ground plane of the scene. Thus, we can approximate the up direction in camera coordinates by taking the normal plane containing all three cameras. In order to resolve uncertainty about whether this is the up direction or its negation, we check that the cosine similarity with the camera directions is negative, as they lie in the upper halfspace in world coordinates. Given a camera center in camera coordinates, we can thus create a camera pose. To define these centers, we uniformly sample a circle in the plane containing all three cameras, which approximately goes through each one (the radius is the mean distance from the approximate world origin). We choose $K=50$ camera centers and create camera poses with the estimated world origin and up direction for the finetuning process. We use these to approximate \cref{eq.finetune}.

\begin{figure*}
    \centering
    \includegraphics[width=\linewidth]{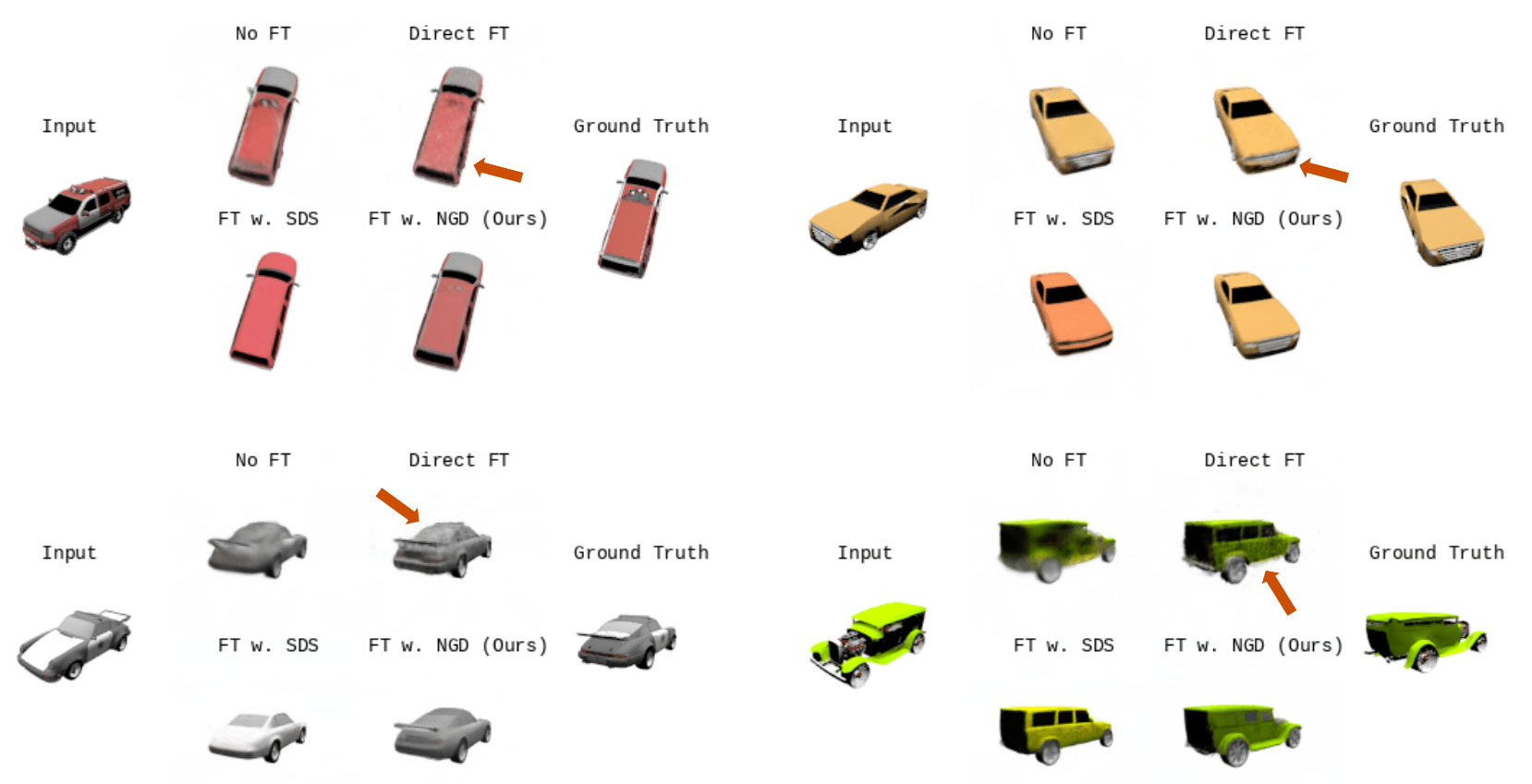}
    \caption{Qualitative examples of ablation studies on fine-tuning strategy. FT refers to finetuning. The red arrow shows floating and noisy artifacts due to learning from inconsistent CDM predictions.}
    \label{fig:ablation_sample}
\end{figure*}

\section{Details of Baseline Methods}
The baselines are shown in Table.~\ref{tab:results}, we gathered the error metrics of LFN~\cite{Sitzmann2021LightRendering}, 3DiM~\cite{watson2022novel} on the ShapeNet dataset from their respective papers. 
As for SRN~\cite{srns2}, PixelNeRF~\cite{pixelnerf}, CodeNeRF~\cite{codenerf}, FE-NVS~\cite{guo2022fast}, and VisionNeRF~\cite{visionnerf}, we obtain the ShapeNet results from the VisionNeRF paper and conduct FID calculation using the renderings provided by the authors of PixelNeRF and VisionNeRF.
Moreover, to compare against VisionNeRF on the ABO and Clevr3D datasets, we used its publicly available source code and modified the dataloader accordingly.
For training, we use the same hyperparameter setup denoted by the VisionNeRF paper.
Namely, we set the image feature channels to 512.
The learning rate of the feature extractor is set to $1e{-5}$ and MLP to $1e{-4}$.
We keep the same learning rate schedule and apply warm-up and decay, as shown in the original paper.
We trained VisionNeRF for 500k steps on the ABO dataset and 250k steps on the Clevr3D dataset since we found it easier to overfit the Clevr3D scenes.
Moreover, we also adjusted the batch and ray bundle sizes to fit into the GPU memory.

\section{Additional Comparison Details}\label{section:comparisondetails}
\subsection{Importance of 3D-aware Diffusion}
In \cref{table:3dawarecdm}, we showed the comparison with different CDM architectures. ``Concat'' means directly concat the input view with the noisy image, while ``Cross-attention'' adopts a similar conditioning as X-UNet~\cite{watson2022novel}. Both did not involve the volume rendering in the encoding time, which can be seen as 3D-unaware. The results showed that, when applying a 3D-aware diffusion model, we can consistently achieve better results and generate more coherent views based on the input.

\begin{table}[h]
\centering
\small
\begin{tabular}{lcccc}
\toprule
Method & PSNR$\uparrow$ & SSIM$\uparrow$ & LPIPS$\downarrow$ & FID$\downarrow$\\
\midrule
No CDM Fine-tuning &{23.81} & 0.915 & 0.093 & 42.37\\
\midrule
CDM architecture \\
\; \; Concat          & 20.72 & 0.874 & 0.135 & 56.27 \\
\; \; Cross-attention & 21.13 & 0.885 & 0.123 & 35.49\\
\; \; 3D-aware (Ours) & {\bf 23.51} & {\bf 0.917} & {\bf 0.082} & {\bf 18.09}\\
\bottomrule
\end{tabular}
\caption{Experiments of showing importance for 3D-aware CDM on ShapeNet-Cars.} 
\label{table:3dawarecdm}  
\end{table}

\subsection{Fine-tuning Strategies}
In \cref{table:finetuneresults} and \cref{fig:ablation_sample}, we showed results for multiple sampling methods for finetuning NeRF. Here we give the details of these methods. 

\paragraph{v.s. Direct distillation}
For Direct distillation, we directly sample virtual views from the CDM given the initial pixelNeRF renderings and use these to finetune the NeRF directly with a standard L2 loss. Note that these renderings are unlikely to be multiview consistent as the denoising process takes place independently for each. Thus, the resultant renderings are inconsistent and incongruous with the input, which is also reflected in the learned NeRF.

\paragraph{v.s. Score distillation sampling (SDS)}
We also compared SDS~\cite{poole2022dreamfusion}, where virtual views are continually predicted by adding noise directly to renderings and taking an L2 loss between the NeRF rendering and the resultant denoised images. Here we note three significant differences with SDS, which may result in its poorer performance:

\begin{enumerate}
    \item Inconsistent noise schedule. In our method, we only sample once per view, continually decreasing the noise with greater NeRF guidance. In contrast, SDS will provide inconsistent gradient updates as random amounts of noise are added to the NeRF renderings and then denoised, yielding blurry results which regress the mean of the supervision.
    \item The learned score function of the CDM may be inadequate. That is to say, the modes of the PDF may not reflect sharp images of the dataset, causing poorer results. Our method uses NeRF to guide the process, which avoids directly seeking a mode.
    \item Out-of-distribution inputs. At low noise levels, rendering a NeRF with additional noise will not resemble a real image with a similar amount of noise. Thus, the inputs to the denoiser may be out-of-distribution. In contrast, our method uses the CDM to refine the NeRF during sampling, keeping the samples close to the data manifold.
\end{enumerate}

\paragraph{v.s. Stochastic Conditioning}
As shown in the main paper, naively finetuning the NeRF parameters from the CDM's generation typically leads to noisy results with severe floating artifacts. It is the inconsistent CDM predictions that cause conflicts in learning NeRF. 
Targeting on this, \citet{watson2022novel} proposed ``stochastic conditioning'' -- an autoregressive approach for synthesizing virtual views in a sequence, where for generating a novel view, each diffusion step stochastically conditions on previously generated views. Although this model's dependencies are across the virtual views, the generated images are not guaranteed to be multiview consistent. 
Moreover, our initial exploration showed that the imperfect autoregressive prediction accumulated errors easily, resulting in degenerated results for long sequences without stable geometry.

\section{Additional Qualitative Results}
\label{section:additional_results}
\begin{figure*}
    \centering
    \includegraphics[width=0.86\linewidth]{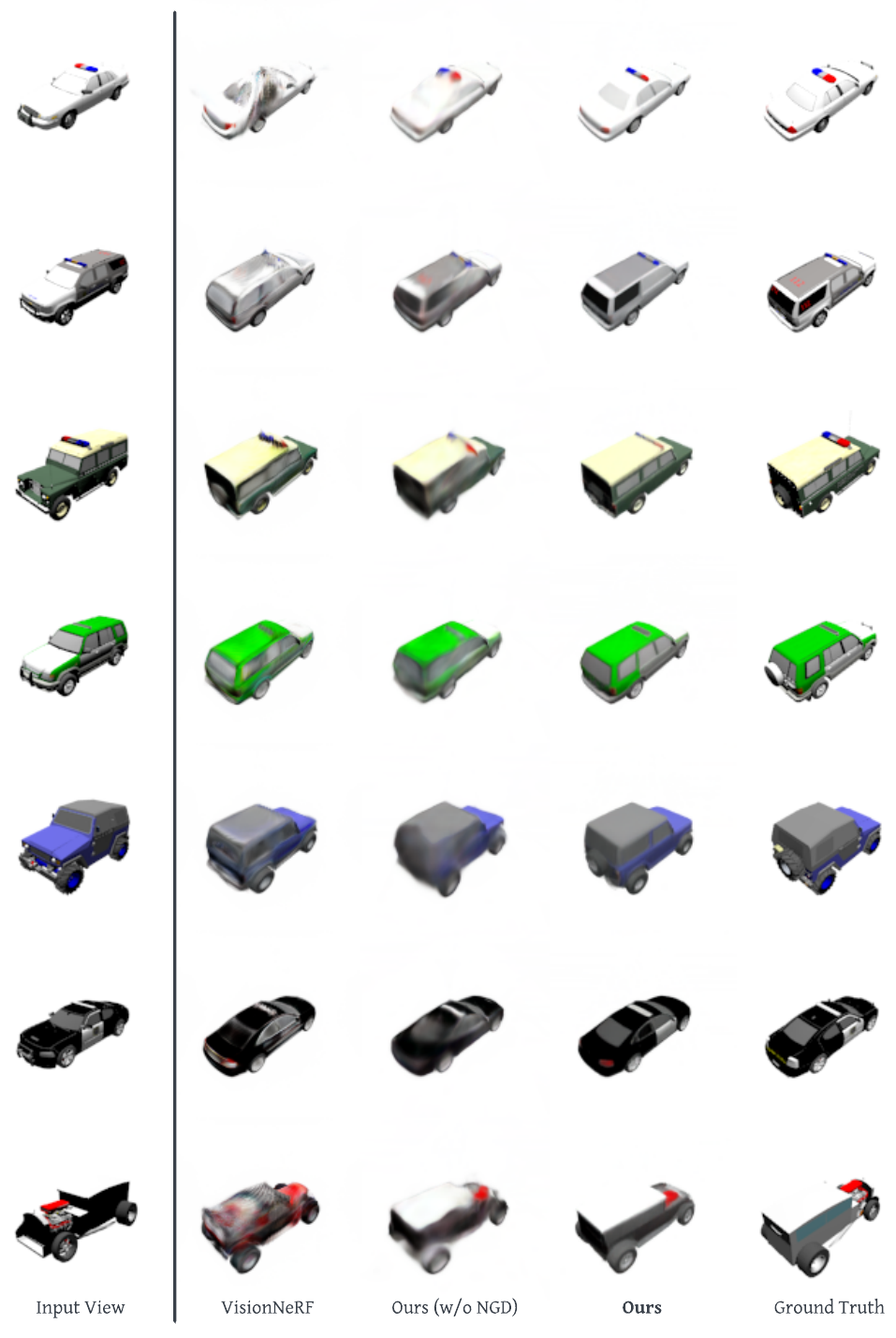}
    \caption{Additional examples of single-image view synthesis on ShapeNet Cars.}
    \label{fig:additional_cars}
\end{figure*}

\begin{figure*}
    \centering
    \includegraphics[width=0.87\linewidth]{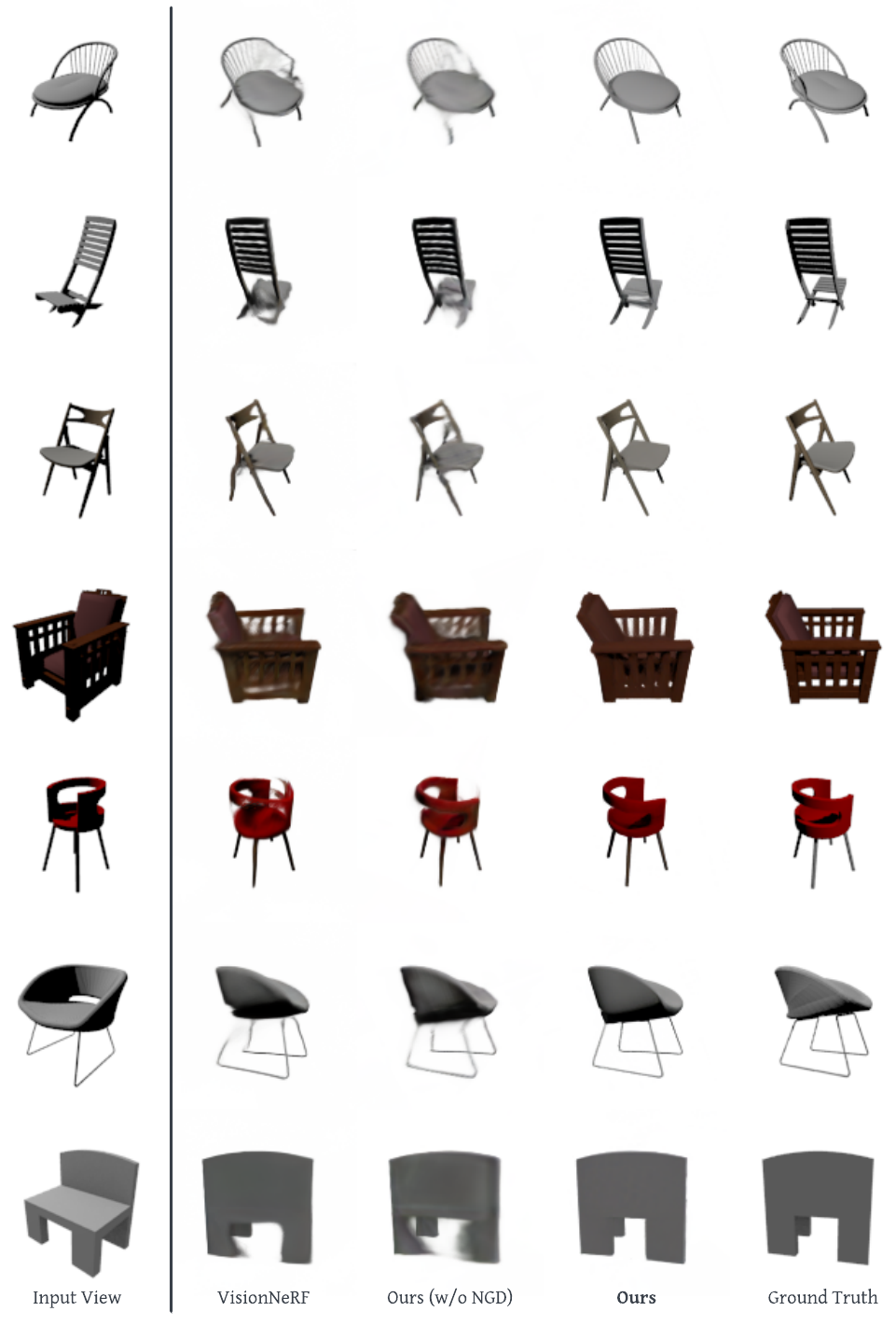}
    \caption{Additional examples of single-image view synthesis on ShapeNet Chairs.}
    \label{fig:additional_chairs}
\end{figure*}

\begin{figure*}
    \centering
    \includegraphics[width=0.87\linewidth]{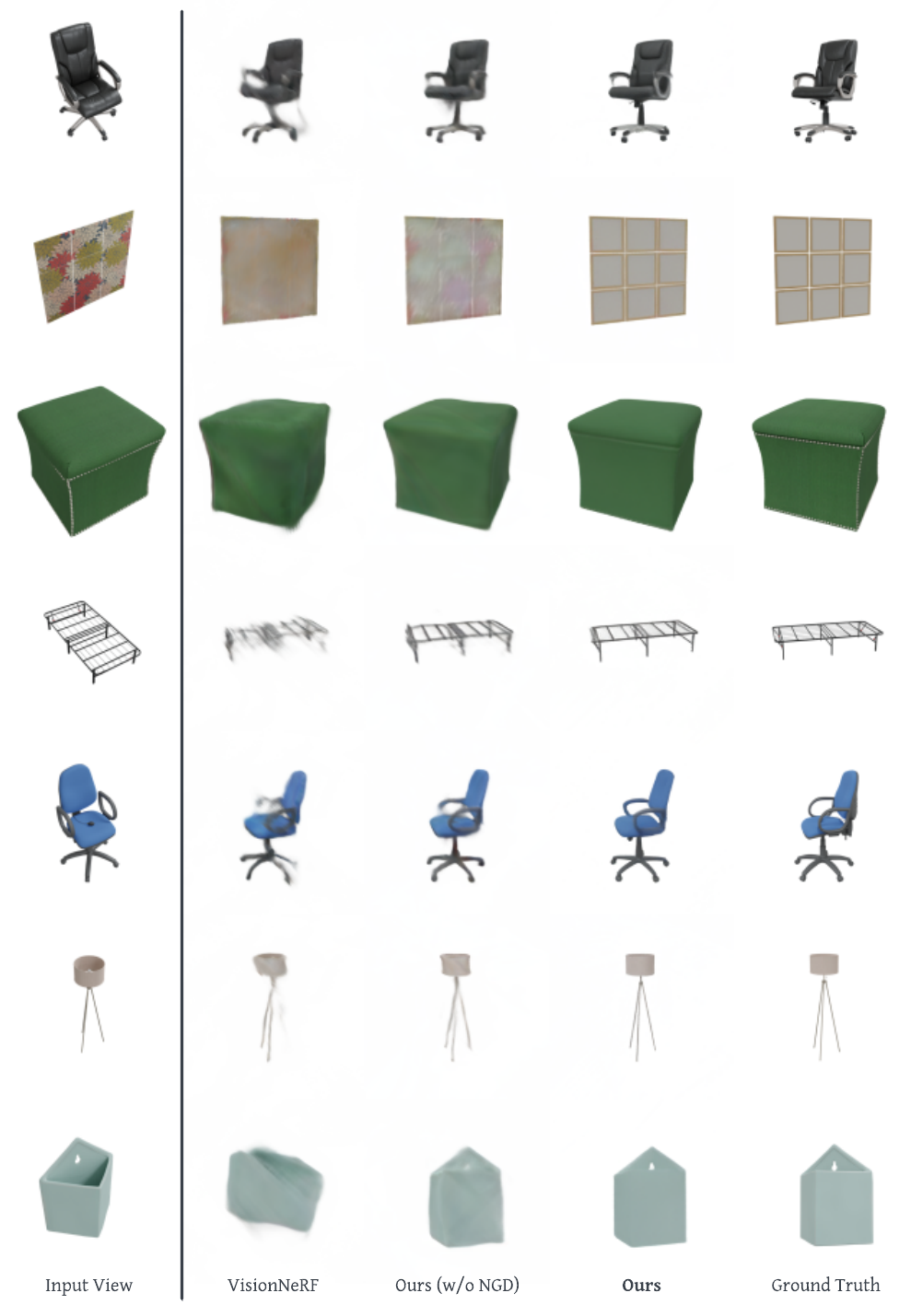}
    \caption{Additional examples of single-image view synthesis on ABO dataset.}
    \label{fig:additional_abo1}
\end{figure*}

\begin{figure*}
    \centering
    \includegraphics[width=0.87\linewidth]{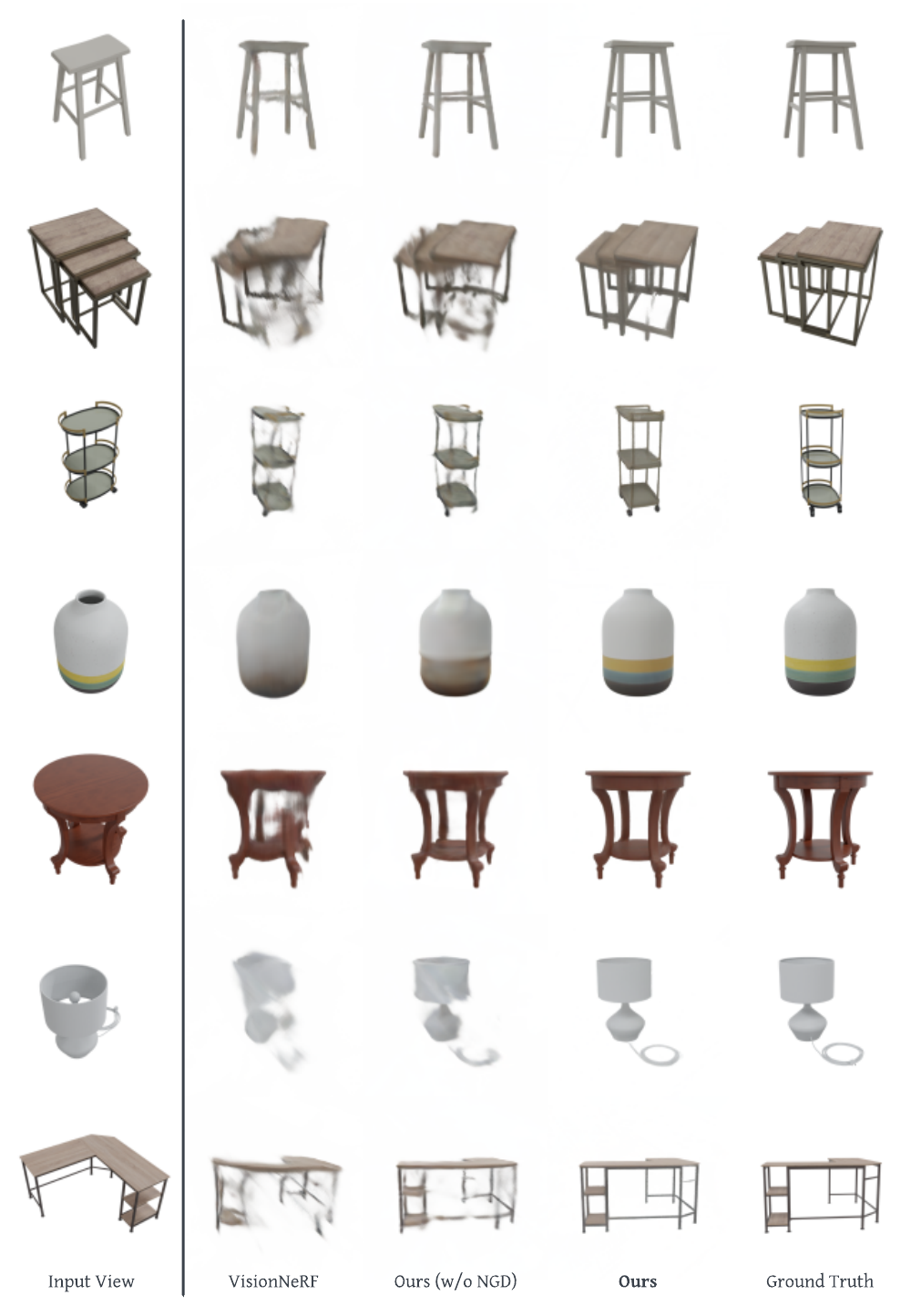}
    \caption{Additional examples of single-image view synthesis on ABO dataset.}
    \label{fig:additional_abo2}
\end{figure*}

Finally, we provide additional qualitative results for our base single-image models compared with VisionNeRF~\cite{visionnerf} on ShapeNet Cars~(\cref{fig:additional_cars}), Chairs~(\cref{fig:additional_chairs}) and ABO dataset~(\cref{fig:additional_abo1,fig:additional_abo2}). Images are rendered at a specific viewpoint given a single image input.
Please refer to supplementary materials for more video results.



\end{document}